\pdfoutput=1
\documentclass{article}

\PassOptionsToPackage{numbers}{natbib}
 \usepackage[main,final]{neurips_2026}

\usepackage{pifont}
\usepackage{graphicx}
\usepackage{algorithm}
\usepackage{algpseudocode}

\newcommand{\cmark}{\ding{51}}
\newcommand{\xmark}{\ding{55}}
\usepackage{multirow}
\usepackage{amssymb}
\usepackage{amsmath}
\usepackage[utf8]{inputenc} %
\usepackage[T1]{fontenc}    %
\usepackage{hyperref}       %
\usepackage{url}            %
\usepackage{booktabs}       %
\usepackage{amsfonts}       %
\usepackage{nicefrac}       %
\usepackage{microtype}      %
\usepackage{xcolor}         %

\title{Unlocking Any-Order Generation in Pretrained Autoregressive Image Models}

\author{%
  Rishav Pramanik\textsuperscript{1}~~~~Marco Pedersoli\textsuperscript{2,3}~~~~Zhaozheng Yin\textsuperscript{1}\\
  \textsuperscript{1}Stony Brook University, New York, USA\\
  \textsuperscript{2}International Laboratory on Learning Systems (ILLS)\\
  \textsuperscript{3}LIVIA, \'{E}TS Montr\'{e}al, QC, Canada\\
}

\begin{document}

\maketitle

\begin{abstract}

Autoregressive (AR) image generators are powerful in principle but inflexible in practice: training with a fixed raster-scan order hardcodes a single factorization of the image distribution into the model, leaving it unable to inpaint, outpaint, or edit at inference time. Existing solutions, including the modern discrete-diffusion family, train from scratch on a mixture of token orderings, paying roughly $3\times$ the compute of standard raster-scan training; discrete diffusion further sacrifices compatibility with the AR inference ecosystem (KV caching, vLLM) by construction. We argue that this cost is unnecessary. A pretrained raster-scan AR model has already learned the joint distribution over image tokens; what it lacks is not knowledge but the ability to express that knowledge through non-sequential conditional pathways. We propose a recipe for adapting pretrained raster-scan models into any-order generators using two components: \textit{future-aware positional embeddings} that resolve a target-position identifiability problem at negligible overhead, and an \textit{ordering curriculum} that traces a stable optimization path from the pretrained solution to a good any-order basin. Applied to four base models across two architectures and two scales (LlamaGen-L/XL, RAR-L/XL), the recipe substantially outperforms MaskGIT on ImageNet inpainting (FID $6.21$ vs.\ $10.44$) and outpainting, is competitive with MAR-L, trades a $1.2$--$1.9$ increase in class-conditional FID for this flexibility, and costs $13$--$33\%$ additional compute beyond pretraining, while leaving the standard AR inference stack untouched. Inpainting, outpainting, and editing, our results suggest, need not be added to autoregressive image models. They can be unlocked from the ones that exist. Code is available at \url{https://github.com/Ordered-AR/unlockedadaptation}.

\end{abstract}

\section{Introduction}
\label{sec:intro}
\begin{figure}[ht!]
    \centering
    \includegraphics[width=\linewidth,keepaspectratio]{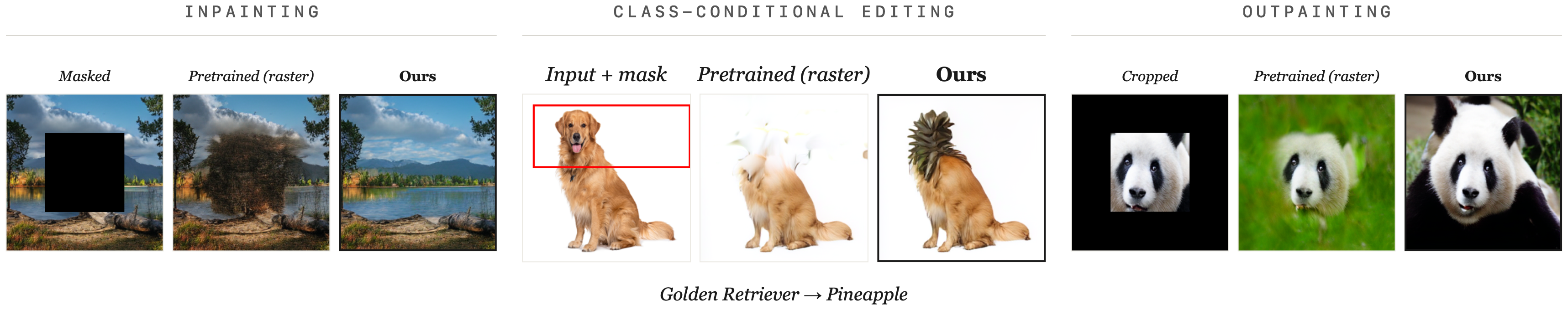}
    \caption{A single autoregressive image model, adapted from a pretrained raster-scan generator, performs inpainting (left), region-conditioned editing (center), and outpainting (right) without task-specific training, sampler modification, or auxiliary networks. The ``Pretrained (raster)'' column shows the same LlamaGen-L~\cite{sun2024autoregressive} model from which our model is adapted, prompted to perform the same tasks under the natural raster-scan generation procedure. Failure modes in this column reflect the structural incapacity of fixed-order AR models to condition on non-contiguous context, not weak performance at a shared task. After adaptation (``Ours''), the same weights produce coherent results across all three capabilities, varying only in which tokens are observed at inference time. Quantitative comparisons against MaskGIT~\cite{chang2022maskgit} on inpainting and outpainting are reported in Section~\ref{sec:results}.}
    \label{fig:teaser}
\end{figure}
Autoregressive (AR) modeling has become a common paradigm for image generation~\cite{li2024autoregressive,sun2024autoregressive,tian2024visual,yu2025randomized}, inheriting the scalability and training stability~\cite{kaplan2020scaling,henighan2020scaling,hoffmann2022an,borgeaud2022improving} that made it standard in language~\cite{carlini2026mythos,openai2026gpt53codex,qwen36plus,team2023gemini}. By tokenizing images into discrete codes and predicting them sequentially~\cite{esser2021taming,sun2024autoregressive,ramesh2021zero,wu2025janus}, AR models reach generation quality comparable to that of leading diffusion and bi-directional models, while sharing the architecture of their language counterparts. Standard AR image models, however, impose a fixed generation order, typically raster scan, proceeding left to right and top to bottom. This ordering is convenient, but it hardcodes a single factorization of the image distribution into the model, leaving it unable to condition on non-contiguous context at inference time.

\textit{Any-order generation} removes this restriction. Rather than committing to a single token ordering, an any-order AR model predicts any subset of image tokens conditioned on any other subset. This unlocks a range of capabilities from a single model: by treating observed tokens as context and sampling the remainder, the same weights perform inpainting, outpainting, and region-conditioned editing without inference-time modification, and admit parallel decoding strategies that predict multiple unobserved tokens jointly rather than strictly one at a time~\cite{pramanik2026distillingspecializedordersvisual,pang2025randar}. Diffusion models support analogous tasks, but only through specialized samplers~\cite{zhang2024hive,lugmayr2022repaint} or auxiliary conditioning networks~\cite{benarous2025spie,meng2022sdedit} layered onto a fixed generation process. Any-order AR offers them as \emph{zero-shot capabilities} of the native generation procedure: one model, one set of weights, one decoding loop.

The obstacle is cost of training. Existing approaches to any-order image generation train from scratch on a mixture of token orderings, a joint optimization that takes roughly $3\times$ the compute of raster-scan training (e.g., RandAR~\citep{pang2025randar} requires 300 epochs vs.\ 100 for the corresponding raster-scan LlamaGen-XL~\citep{sun2024autoregressive}). The most common modern approach is \textit{discrete diffusion}~\citep{chang2022maskgit,weber2024maskbit,yu2025frequency}, which re-frames any-order generation as iterative mask-token denoising under a non-causal transformer. Discrete diffusion reaches competitive quality, but its bidirectional attention and iterative-refinement decoding place it outside the standard AR inference ecosystem: KV caching, speculative decoding, and AR-oriented serving frameworks such as vLLM~\citep{kwon2023efficient} do not accelerate it. The result is not just a training cost gap but a recurring inference gap: practitioners with existing AR pretraining pipelines and AR serving infrastructure cannot adopt these methods without parallel investment in a separate stack.

We argue that this cost reflects a conceptual conflation rather than a fundamental requirement. \textit{Any-order AR models need not be trained from scratch; they can be adapted from existing pretrained raster-scan models.} A raster-scan AR model has already done the expensive work: it has learned an approximation of the joint distribution over image tokens, encoding texture statistics, spatial coherence, and semantic structure into its representations. The chain rule of probability guarantees that the \textit{true} joint $p(x_1, \ldots, x_n)$ factorizes equivalently under any token ordering, so a model that is a good fit under one factorization is a candidate fit under all of them. We hypothesize that the representations learned under raster-scan training are sufficiently order-agnostic to support any-order generation with light adaptation, because they encode properties of images rather than properties of scan lines. What the model lacks is not distributional knowledge but the ability to express that knowledge through non-sequential conditional pathways, and acquiring that ability, starting from a good initialization, need not be expensive.

To test this hypothesis, we propose a recipe for adapting a pretrained raster-scan AR model into an any-order generator. Two components are jointly necessary. First, we introduce \textit{future-aware positional embeddings} that inject the identity of the next target position into each step of the generation sequence. This addresses an identifiability problem inherent to causal any-order generation: without target-position information, identical context must produce different outputs for different targets, generating conflicting gradient signals that prevent convergence. Our embeddings address this at negligible overhead, preserving the standard single-stream transformer architecture and the inference-time properties it confers. Second, we employ an \textit{ordering curriculum} that gradually interpolates from raster-scan to fully random permutations over the course of adaptation. Without it, sudden exposure to random orderings produces large, high-variance gradients that move the model out of the pretrained basin before order-flexible routing has developed, eroding the representations we aim to reuse. The curriculum traces a stable optimization path from the pretrained solution to a good any-order basin. Both components are necessary: ablating either causes training to diverge or generation quality to drop noticeably (Section~\ref{sec:results}).

The resulting model reaches any-order generation at roughly $2.4\times$ lower end-to-end cost than training from scratch, requiring only $25\%$ additional computation beyond the original raster-scan pretraining budget on LlamaGen-XL ($13$--$33\%$ across base models); for practitioners who already have a pretrained AR model, the marginal adaptation cost is over an order of magnitude lower ($12\times$) than training any-order from scratch. On LlamaGen-XL, FID moves from $2.63$ to $3.91$ while Inception Score, Precision, and Recall remain close to the pretrained values ($252.94$, $0.81$, $0.53$), a measured quality-flexibility trade-off rather than a lossless conversion. We observe a similar pattern across LlamaGen-L and RAR-L/XL (Section~\ref{sec:results}). The adapted model also gains the any-order capabilities promised by the construction (inpainting, outpainting, and region-conditioned editing), where it substantially outperforms MaskGIT~\cite{chang2022maskgit} on ImageNet inpainting (FID $6.21$ vs.\ $10.44$) and outpainting (FID $5.84$ vs.\ $12.17$) and is competitive with MAR~\cite{li2024autoregressive}, despite never having been trained on either task (Section~\ref{sec:results}). Because the recipe leaves the causal transformer architecture untouched, the adapted model retains KV caching and parallel decoding and can be served with vLLM~\cite{kwon2023efficient} through a custom model registration, whereas discrete-diffusion any-order approaches cannot reuse the KV cache, by construction. Flexible image generation (inpainting, outpainting, editing) need not be added to autoregressive models. It is, in our results, already accessible from the ones that exist.

\section{Related Work}

\paragraph{Autoregressive image generation.}
Autoregressive image models, long standard in language modeling, have recently reached generation quality on par with leading diffusion methods on standard benchmarks~\cite{li2024autoregressive,sun2024autoregressive,tian2024visual,ren2025beyond}. They tokenize images via a quantized autoencoder~\cite{esser2021taming,sun2024autoregressive,tian2024visual,yu2025randomized} and predict discrete tokens with a transformer, inheriting the inference ecosystem developed for LLMs (KV caching, paged attention, tensor/pipeline parallelism)~\cite{kwon2023efficient} and integrating naturally into unified multimodal stacks~\cite{wu2025janus,deng2025emerging}. Standard AR image models commit to a fixed raster-scan order; this work relaxes that constraint while preserving the inference-time properties that make AR attractive.

\paragraph{Any-order generation.}
Permutation language modeling~\cite{yang2019xlnet} showed that a single causal model can generate under arbitrary orderings if a target-position identifiability problem is resolved, originally via two-stream attention. Image-domain follow-ups~\cite{pang2025randar,yu2024randomized,hoogeboom2022autoregressive} apply the same recipe to discrete image tokens but pay similar costs: any-order capability obtained from scratch, with architectural overhead or training budgets several times that of standard raster-scan AR. MAR~\cite{li2024autoregressive} instead predicts continuous tokens with a diffusion head, also from scratch. The current dominant alternative is \emph{discrete diffusion}~\cite{chang2022maskgit,weber2024maskbit,chang2023muse}, which casts any-order generation as iterative mask-token denoising under bidirectional attention but, by construction, sits outside the AR inference ecosystem. We instead retain causal AR, address identifiability with a lightweight additive embedding, and obtain any-order capability through adaptation rather than from-scratch training.

\paragraph{Adaptation of pretrained models.}
Adapting pretrained models to new objectives is standard practice in language and vision~\cite{ke2022continual,liu2023visual,hu2022lora,han2024parameterefficient}. Most prior work targets new \emph{tasks} or \emph{domains}; we instead target a new \emph{factorization} of the same data distribution. Task-specific methods such as ControlNet~\cite{zhang2023adding} are complementary: they add a branch for one fixed control signal, whereas we support arbitrary observed-token subsets. Whether representations learned under one factorization transfer to others is an empirical question; our results suggest that they do, with light adaptation.

\section{Method}
\label{sec:method}
\label{sec:motivation}

We consider an autoregressive image model that predicts a sequence of discrete tokens $x = (x_1, \ldots, x_n)$ produced by a tokenizer applied to an image. A standard AR model factorizes the joint under a fixed raster-scan order $\sigma_{\mathrm{raster}} = (1, \ldots, n)$. An any-order model relaxes this by allowing $\sigma$ to vary, and is trained to maximize the expected log-likelihood under a distribution over permutations:
\begin{equation}
\mathcal{L}_{\mathrm{any\text{-}order}}(\theta) = \mathbb{E}_{\sigma \sim P}\left[\, \sum_{t=1}^{n} \log p_\theta\bigl(x_{\sigma(t)} \,\big|\, x_{\sigma(<t)}\bigr) \,\right],
\label{eq:any-order-objective}
\end{equation}
with $P = \mathrm{Uniform}(S_n)$ for full any-order training. By the chain rule of probability, every choice of $\sigma$ is a valid factorization of the same joint. The question we address is whether a model pretrained under raster scan can be adapted, by continued training under Equation~\ref{eq:any-order-objective}, into one that supports the full family of factorizations. We argue that this is possible, but only if two distinct issues are addressed.

\paragraph{The identifiability problem.}
\label{sec:identifiability}
Under raster-scan, the prediction target at step $t$ is implicitly $\sigma(t+1)=t+1$ and the model never needs to be told \emph{which} position to predict. Under varying $\sigma$, that contract breaks: the same observed context $x_{\sigma(<t)}$ corresponds to different prediction targets across permutations. Concretely, predicting a sky token given the rest of an image is a different problem from predicting a face token given the same context: the visible information is identical, but the conditional distributions are not. A model that attempts to fit $p(x_j \mid x_S)$ and $p(x_k \mid x_S)$ for $j \neq k$ from the same input representation therefore receives conflicting gradients, and the objective in Equation~\ref{eq:any-order-objective} becomes ill-posed without target-position information. XLNet~\citep{yang2019xlnet} resolved this in language with a two-stream attention mechanism that maintains a separate query stream for target-position information; this restores identifiability but doubles per-layer attention compute and breaks compatibility with KV caching. We propose an alternative based on input-level conditioning (Section~\ref{sec:method-embeddings}).

\paragraph{The optimization problem from a pretrained initialization.}
\label{sec:curriculum-motivation}
Even with identifiability resolved, naive any-order training from the pretrained parameters $\theta^*$ fails. At $\theta^*$, the per-permutation loss $\mathcal{L}_\sigma$ is small for $\sigma \approx \sigma_{\mathrm{raster}}$ but large for random $\sigma$, and an expectation over $\mathrm{Uniform}(S_n)$ is therefore dominated by high-loss random permutations. The resulting gradients are large in magnitude and inconsistent in direction across permutations; applied at $\theta^*$, they push the model out of the pretrained basin before order-flexible routing has developed, a form of catastrophic forgetting that erodes the very representations we aim to reuse. We address this with an ordering curriculum (Section~\ref{sec:method-curriculum}) that interpolates from raster-scan to uniform random over the course of adaptation, tracing a stable optimization path from $\theta^*$ to a good any-order basin. Note that this is distinct from simply lowering the learning rate: a smaller step size leaves the gradient \emph{direction} unchanged and continues to point away from the pretrained basin. The curriculum instead modifies the \emph{composition} of the loss, weighting raster-like permutations heavily early on so that the gradient itself is dominated by directions consistent with the pretrained solution. Appendix~\ref{app:motivation-extended} gives the full gradient-magnitude and gradient-variance argument and frames it as a continuation method. The two mechanisms address orthogonal aspects of the problem and are jointly necessary: identifiability makes the loss well-posed; the curriculum makes the well-posed loss reachable from the pretrained initialization. Section~\ref{sec:ablations} confirms both halves empirically.

\subsection{Future-Aware Positional Embeddings}
\label{sec:method-embeddings}

We introduce a learned embedding table $E \in \mathbb{R}^{n \times d}$, where $d$ is the model's hidden dimension and row $E_j$ encodes the identity of position $j$ as a target. Under permutation $\sigma$, the input representation at sequence position $t$ is augmented with the embedding of the \emph{next} target position:
\begin{equation}
\tilde{h}_t = h_t + E_{\sigma(t+1)},
\label{eq:future-aware}
\end{equation}
where $h_t = e(x_{\sigma(t)})$ is the standard input embedding of the token at sequence position $t$ (for RAR, including its absolute position embedding at $\sigma(t)$). The final token has no successor; we use a learned null embedding $E_{\varnothing}$. The embedding $E$ is shared across sequence positions and permutations: the same row $E_j$ is added at any sequence position whose successor under $\sigma$ is the spatial position $j$. The addition is at the input layer only; we considered AdaLN-style injection at every layer~\citep{peebles2023scalable} but found a single input-level addition sufficient, consistent with the view that the embedding's role is to disambiguate the prediction target rather than to participate in feature computation throughout the network.

\paragraph{Position handling under permutation.}
The two base-model families we adapt encode position differently (LlamaGen via rotary position embeddings (RoPE)~\citep{su2024roformer} and RAR via additive learned absolute position tables at the input layer), but the principle for any-order generation is the same in both: at sequence step $t$, the position used for rotation (LlamaGen) or table look-up (RAR) is the spatial index $\sigma(t)$, not the sequence step. With this choice, attention or input addition between tokens at sequence steps $t$ and $t'$ depends only on their spatial positions $\sigma(t)$ and $\sigma(t')$, identical to what the model would have computed under raster scan with those tokens at the same spatial locations. The model thus inherits the spatial-relational priors learned during pretraining unchanged; only the order in which sequence steps consume those positions varies.

Compared to two-stream attention~\citep{yang2019xlnet}, the additive embedding adds only $(n+1) \cdot d$ parameters and an elementwise sum, leaves the standard single-stream transformer architecture untouched, and preserves KV caching and parallel decoding.

\subsection{Ordering Curriculum}
\label{sec:method-curriculum}

We sample training permutations from a mixture parametrized by a curriculum coefficient $\alpha \in [0, 1]$:
\begin{equation}
\sigma \sim P_\alpha = (1 - \alpha) \cdot \delta_{\sigma_{\mathrm{raster}}} + \alpha \cdot \mathrm{Uniform}(S_n).
\label{eq:curriculum}
\end{equation}
At $\alpha = 0$, adaptation reduces to continued raster-scan training; at $\alpha = 1$, every example is trained on a uniformly random permutation. We use a linear schedule $\alpha(s) = \min(s / S_{\mathrm{ramp}}, 1)$, holding $\alpha = 1$ after step $S_{\mathrm{ramp}}$. Within a minibatch, each example draws an independent permutation, smoothly interpolating between the raster-scan and any-order objectives as $\alpha$ increases.

\subsection{Training Objective}
\label{sec:method-objective}

The adaptation loss combines the future-aware embedding with the curriculum sampler:
\begin{equation}
\mathcal{L}(\theta) = \mathbb{E}_{\sigma \sim P_{\alpha(s)}}\left[\, \sum_{t=1}^{n} -\log p_\theta\bigl(x_{\sigma(t)} \,\big|\, x_{\sigma(<t)},\, E_{\sigma(t+1)}\bigr) \,\right].
\label{eq:training-objective}
\end{equation}
Apart from the curriculum sampler and the additive embedding, the training pipeline is identical to continued raster-scan AR training. We initialize from the pretrained checkpoint and adapt for $T_{\mathrm{ft}}$ epochs (set per base model so adaptation cost is a small fraction of pretraining). We use the same optimizer and weight decay as pretraining; $S_{\mathrm{ramp}}$ is tuned on a held-out subset. The future-aware embedding table $E$ is zero-initialized (for RAR, its existing target-aware positional embedding is loaded from the checkpoint) and trained at $10\times$ the base learning rate (Section~\ref{sec:ablations}); per-model values are in Appendix~\ref{app:hyperparameters}.

\subsection{Inference}
\label{sec:method-inference}

At inference, the user (or upstream task) specifies a generation order $\sigma$: raster scan recovers the pretrained model's behavior; uniform random shuffles the order; task-induced placements put observed tokens (an inpainting mask, an outpainting boundary, a partial edit) first in $\sigma$ and sample the rest in any chosen order. Generation proceeds sequentially under $\sigma$ with the future-aware embedding $E_{\sigma(t+1)}$ added to each input. Because the architecture is a standard single-stream transformer, KV caching applies unchanged, supporting inpainting, outpainting, and region-conditioned editing as direct instances of conditional generation without task-specific training, sampler modification, or auxiliary networks (Section~\ref{sec:results}).

\section{Results}
\label{sec:results}

We evaluate our adaptation recipe along four axes: (i) generation quality relative to standard generative baselines and the pretrained raster-scan models we adapt (Section~\ref{sec:quality}); (ii) the any-order capabilities the unlocked model acquires, demonstrated on inpainting, outpainting, and region completion (Section~\ref{sec:capabilities}); (iii) the necessity of each recipe component, validating the diagnoses in Section~\ref{sec:method} (Section~\ref{sec:ablations}); and (iv) training and inference cost relative to existing any-order approaches (Section~\ref{sec:compute}).

\subsection{Experimental Setup}
\label{sec:setupres}

\paragraph{Datasets.}
All experiments use ImageNet-1k (ILSVRC-2012)~\cite{russakovsky2015imagenet}, on which all base models and baselines were pretrained. Adaptation uses the $1.28$M-image training split, pre-tokenized with each model's VQ tokenizer. Class-conditional metrics are computed against the standard ImageNet $256 \times 256$ reference statistics~\cite{dhariwal2021diffusion}, inpainting and outpainting are evaluated on the $50{,}000$-image validation split, and region completion on $5{,}000$ held-out images, with masks shared across methods. We use no other datasets.

\paragraph{Base models.}
We apply our recipe to two families of class-conditional autoregressive image generators trained on ImageNet: LlamaGen~\cite{sun2024autoregressive} (sizes L and XL) and RAR~\cite{yu2025randomized} (sizes L and XL). For each base model we start from the publicly released pretrained checkpoint and apply the recipe described in Section~\ref{sec:method}. Image tokens come from the corresponding pretrained VQ tokenizer; we do not modify the tokenizer. LlamaGen-XL generates at $384 \times 384$ following its released configuration and is resized to $256 \times 256$ for evaluation.

\paragraph{Adaptation.}
We adapt each pretrained checkpoint for $T_{\mathrm{ft}}$ epochs (set per base model; see Appendix~\ref{app:hyperparameters}) with the curriculum coefficient $\alpha$ scheduled linearly from $0$ to $1$ over the first $S_{\mathrm{ramp}}$ epochs and held at $1$ for the remainder. Future-aware embeddings are zero-initialized (loaded from the checkpoint for RAR) and trained at $10\times$ the base learning rate; we ablate this choice in Section~\ref{sec:ablations}. All other hyperparameters match the corresponding pretraining configuration.

\paragraph{Evaluation.}
We report Fr\'echet Inception Distance (FID)~\cite{kynkaanniemi2023role}, Inception Score (IS)~\cite{barratt2018note}, and Precision/Recall on $50{,}000$ generated samples following the standard ImageNet protocol. All samples use the same number of decoding steps as the corresponding pretrained model and the published sampling configuration unless otherwise specified.

\subsection{Class-Conditional Image Generation}
\label{sec:quality}

Table~\ref{table:main_results} reports class-conditional generation quality on ImageNet $256 \times 256$, alongside representative diffusion and AR baselines. %
\begin{table}[ht!]
\centering
\caption{Class-conditional generation on ImageNet $256 \times 256$. Our adapted models (rows marked \textbf{Ours}) trade a $1.17$--$1.93$ FID increase relative to the pretrained raster-scan baselines they are adapted from for full any-order capability (Section~\ref{sec:capabilities}). $^*$Reported by the original paper.}
\label{table:main_results}
\small
\setlength{\tabcolsep}{4pt}
\begin{tabular}{c|l|c|c|c|c|c|c}
\toprule
\textbf{Type} & \textbf{Model} & \textbf{\#Para.} & \textbf{FID$\downarrow$} & \textbf{IS$\uparrow$} & \textbf{Precision$\uparrow$} & \textbf{Recall$\uparrow$}  & \textbf{Steps}\\
\midrule
Diffusion & ADM~\cite{dhariwal2021diffusion} & 554M & 4.59 & 186.70 & 0.82 & 0.523 & 250 \\
& LDM-4~\cite{rombach2022high} & 400M & 3.60 & 247.7 & -- & -- & 250\\
& DiT-XL~\cite{peebles2023scalable} & 675M & 2.27 & 278.2 & 0.83 & 0.57 & 250\\
\midrule
Bi-directional AR & MaskGIT-re~\cite{chang2022maskgit} & 227M & 6.80 & 214.0 & 0.82 & 0.51 & 8 \\
& TiTok-S-128~\cite{yu2024image} & 287M & 1.97 & 281.8 & -- & -- & 64 \\
\midrule
Causal AR & VAR~\cite{tian2024visual} & 600M & 2.57 & 302.6 & 0.83 & 0.56 & 10 \\
& VQGAN-re~\cite{esser2021taming} & 1.4B & 5.20 & 280.3 & -- & -- & 256\\
& RAR-B~\cite{yu2025randomized} & 261M & 1.95 & 290.5 & 0.82 & 0.58 & 256 \\
& RAR-L~\cite{yu2025randomized} & 461M & 1.70 & 299.5 & 0.81 & 0.60 & 256 \\
& RAR-XL~\cite{yu2025randomized} & 955M & 1.50 & 306.9 & 0.80 & 0.62 & 256 \\
& LlamaGen-L~\cite{sun2024autoregressivemodelbeatsdiffusion} & 343M & 3.07 & 256.06 & 0.83 & 0.52 & 256 \\
& LlamaGen-XL$^*$~\cite{sun2024autoregressivemodelbeatsdiffusion} & 775M & 2.63 & 244.08 & 0.80 & 0.57 & 576 \\
& RandAR-L~\cite{pang2025randar} & 343M & 2.55 & 288.5 & 0.81 & 0.58 & 88\\
& RandAR-XL~\cite{pang2025randar} & 775M & 2.22 & 314.21 & 0.80 & 0.60 & 256\\
\midrule
\textbf{Ours} & LlamaGen-L + adaptation  & 343M & 4.24 & 231.62 & 0.82 & 0.52 & 256 \\
{\textit{(Causal AR)}}& LlamaGen-XL + adaptation & 775M & 3.91 & 252.94 & 0.81 & 0.53 & 576 \\
& RAR-L + adaptation       & 461M & 3.63 & 283.13 & 0.77 & 0.60 & 256 \\
& RAR-XL + adaptation      & 955M & 2.68 & 285.16 & 0.80 & 0.58 & 256 \\
\bottomrule
\end{tabular}
\end{table}
\begin{figure}
    \centering
    \includegraphics[width=\linewidth,keepaspectratio]{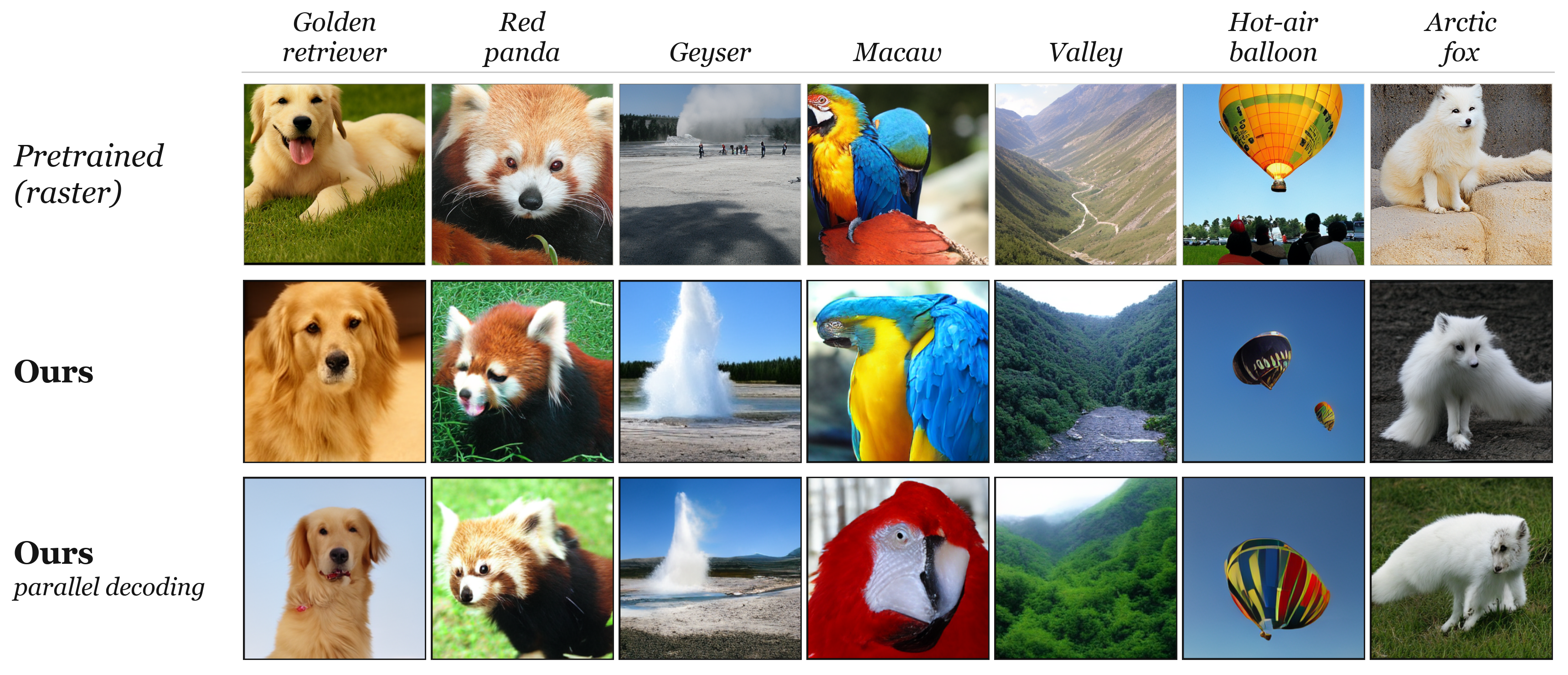}
    \caption{Class-conditional generation after adaptation, further accelerated by parallel decoding. \textbf{Top:} pretrained LlamaGen-L (raster). \textbf{Middle:} our adapted model with sequential decoding. \textbf{Bottom:} our adapted model with parallel decoding (64 steps, $4\times$ fewer forward passes).}
    \label{fig:qualitativeimagenet}
\end{figure}
Adaptation trades some raster-scan quality for any-order capability: across four base models, FID increases by between $1.17$ and $1.93$ relative to the corresponding pretrained checkpoint, while IS, Precision, and Recall remain comparable. We do not claim this cost is fundamental or monotonic with scale. The absolute gap is somewhat larger on RAR (which sits closer to the FID floor, where the same nominal degradation is more visible) than on LlamaGen, but the recipe transfers without modification across both architectures and both scales. This reduction in raster-scan quality (see Fig~\ref{fig:qualitativeimagenet}) buys the full any-order capability suite (Section~\ref{sec:capabilities}) at $\approx 1.25\times$ the original pretraining budget on LlamaGen-XL, against the $\approx 3\times$ required by from-scratch any-order training (Section~\ref{sec:compute}).

\paragraph{Value of pretraining.} At matched total epochs, training the same architecture and curriculum from random initialization is worse than adapting the pretrained checkpoint (Table~\ref{tab:pretraining-value}). This control matches compute but not schedule, since only the pretrained pathway begins with raster-only training.

\begin{table}[ht!]
\centering
\caption{Value of pretraining: FID at matched total epochs with the same architecture and curriculum.}
\label{tab:pretraining-value}
\small
\begin{tabular}{lccc}
\toprule
Model & Total epochs & Pretrained + adapted $\downarrow$ & Random init $\downarrow$ \\
\midrule
LlamaGen-L  & $400$ ($300{+}100$) & $\mathbf{4.24}$ & $5.60$ \\
LlamaGen-XL & $125$ ($100{+}25$)  & $\mathbf{3.91}$ & $4.81$ \\
\bottomrule
\end{tabular}
\end{table}

\subsection{Any-Order Capabilities: Inpainting, Outpainting, and Editing}
\label{sec:capabilities}
\begin{figure}[t]
\centering
\begin{minipage}[t]{0.47\linewidth}
\vspace{0pt}\centering
\includegraphics[width=\linewidth,keepaspectratio]{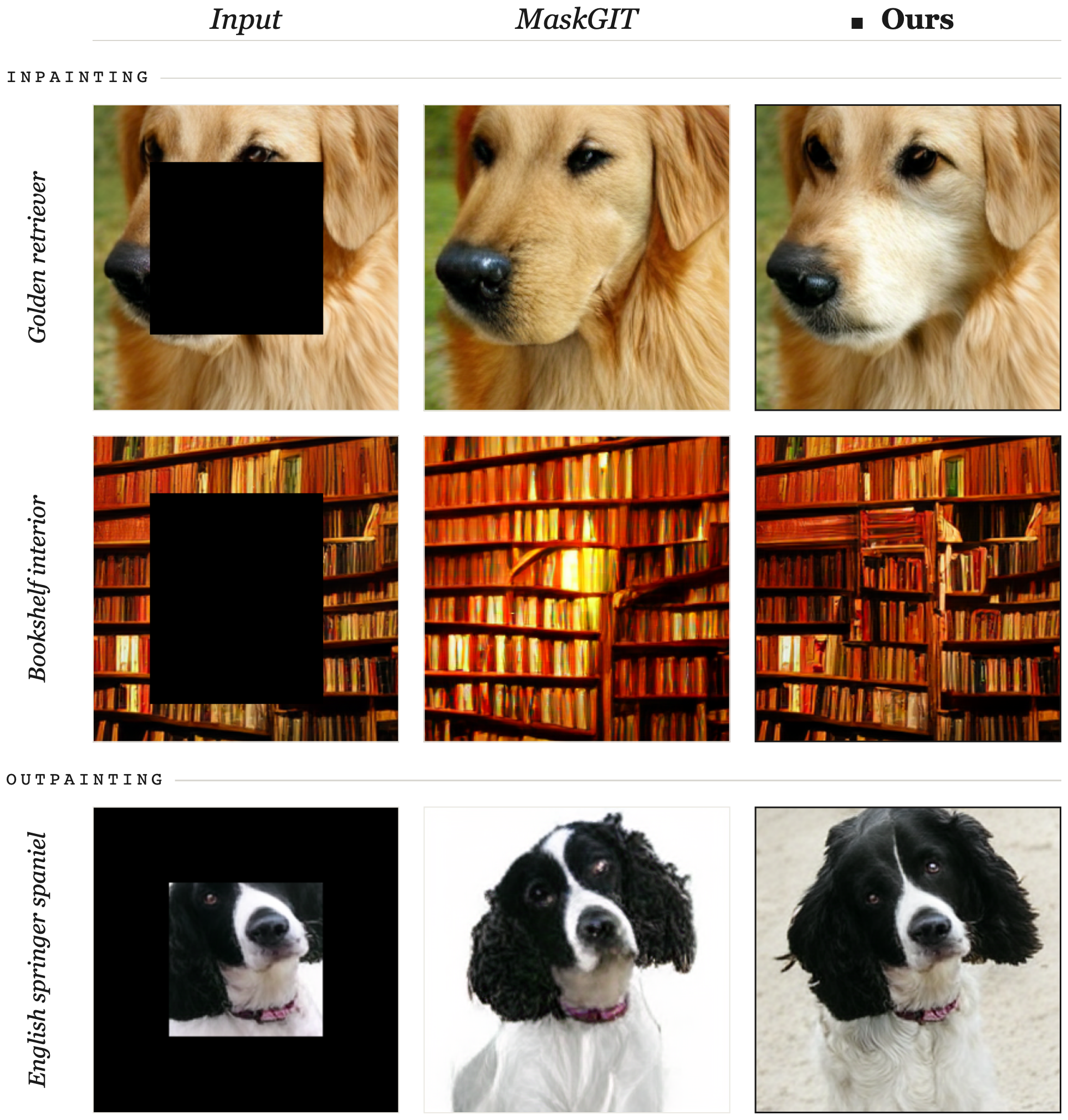}
\caption{Qualitative comparison on inpainting (top) and outpainting (bottom) against MaskGIT.}
\label{fig:capabilities-grid}
\end{minipage}\hfill
\begin{minipage}[t]{0.49\linewidth}
\vspace{0pt}\centering
\includegraphics[width=\linewidth,keepaspectratio]{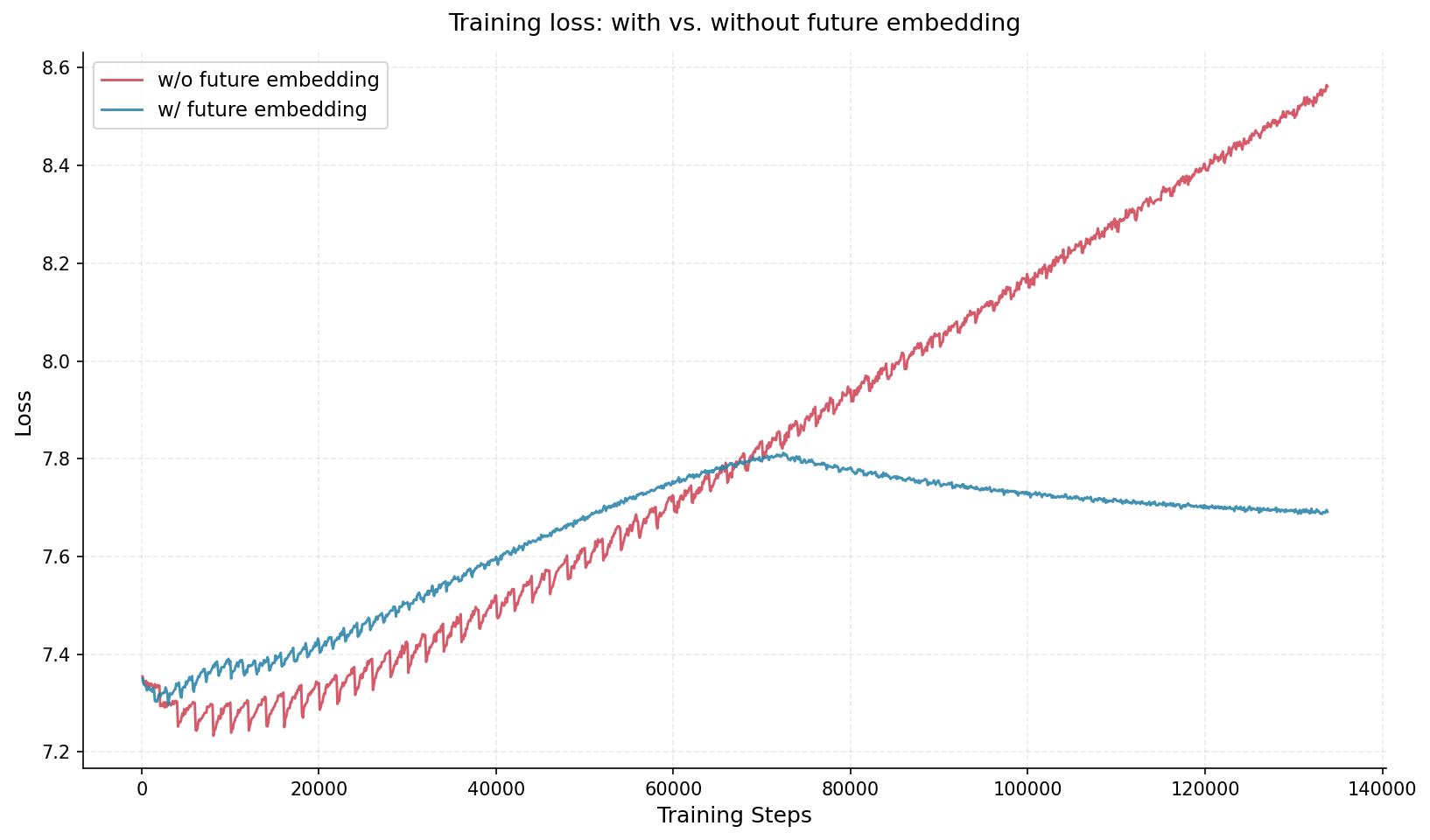}
\caption{Training loss under our full recipe and the future-aware embedding ablation. \textbf{Without future-aware embeddings} (red), the loss fails to descend and instead climbs as the curriculum ramps in random orderings, reflecting the identifiability conflict diagnosed in Section~\ref{sec:method}. The full recipe (blue) descends smoothly throughout.}
\label{fig:loss-curves}
\end{minipage}
\end{figure}
A model that can generate in any order can perform inpainting and outpainting as direct instances of conditional generation under task-induced orderings (Section~\ref{sec:method-inference}). We evaluate both tasks against MaskGIT~\cite{chang2022maskgit}, with its original cosine schedule and with the Halton schedule~\cite{besnier2025halton}, and against MAR-L~\cite{li2024autoregressive}, using released ImageNet checkpoints, the same evaluation protocol, and the same masks across methods.

\begin{table}[ht!]
\centering
\caption{Inpainting and outpainting on ImageNet $256 \times 256$, evaluated on $50{,}000$ samples with TF-Inception. Inpainting uses center masks; outpainting extends a centered $128 \times 128$ crop to full resolution. Lower is better for FID; higher for IS, Precision, Recall.}
\label{tab:capabilities}
\small
\setlength{\tabcolsep}{3.5pt}
\begin{tabular}{@{}l|cccc|cccc@{}}
\toprule
& \multicolumn{4}{c|}{Inpainting} & \multicolumn{4}{c}{Outpainting} \\
Method & FID $\downarrow$ & IS $\uparrow$ & Prec.\ $\uparrow$ & Rec.\ $\uparrow$ & FID $\downarrow$ & IS $\uparrow$ & Prec.\ $\uparrow$ & Rec.\ $\uparrow$ \\
\midrule
MaskGIT (cosine) & $10.44$ & $116.2$ & $0.680$ & $0.575$ & $12.17$ & $107.0$ & $0.664$ & $0.631$ \\
MaskGIT (Halton) & $7.82$ & $123.5$ & $0.581$ & $0.575$ & $7.91$ & $119.1$ & $0.691$ & $0.612$ \\
MAR-L & $6.63$ & $134.2$ & $0.696$ & $\mathbf{0.598}$ & $6.86$ & $127.2$ & $\mathbf{0.736}$ & $\mathbf{0.649}$ \\
\textbf{Ours} & $\mathbf{6.21}$ & $\mathbf{157.0}$ & $\mathbf{0.710}$ & $\mathbf{0.598}$ & $\mathbf{5.84}$ & $\mathbf{159.4}$ & $0.720$ & $0.637$ \\
\bottomrule
\end{tabular}
\end{table}

Our adapted model outperforms both MaskGIT schedules (Table~\ref{tab:capabilities}), although Halton narrows the gap, so part of the original margin was scheduler-dependent. Against MAR-L, our FID is lower on both tasks while MAR-L has higher outpainting Precision and Recall; without run-level uncertainty, we regard the two as competitive. The qualitative comparison in Figure~\ref{fig:capabilities-grid} bears this out at the per-image level: under inpainting, our method recovers sharper structure and more coherent global lighting than MaskGIT; under outpainting, it synthesises a plausible scene around the cropped subject while preserving identity and palette.

\paragraph{Region-conditioned editing.} On $5{,}000$ held-out ImageNet images, we regenerate a random contiguous region covering $30$--$50\%$ of the image from the rest and measure LPIPS to the original; we also report PSNR and SSIM over the generated region of the center-mask inpainting examples. Our model is best on all three (Table~\ref{tab:region}). Since several completions can be valid, these reference-based metrics complement FID, and they do not measure instruction following or identity preservation, which remain qualitative.

\begin{table}[t]
\centering
\caption{Reference-based completion metrics. LPIPS: random region covering $30$--$50\%$ of $5{,}000$ held-out images. PSNR/SSIM (mean $\pm$ std): generated region of the $50{,}000$ center-mask inpainting examples.}
\label{tab:region}
\small
\begin{tabular}{lccc}
\toprule
Method & LPIPS $\downarrow$ & PSNR $\uparrow$ & SSIM $\uparrow$ \\
\midrule
MaskGIT (cosine) & $0.234$ & $19.20 \pm 3.60$ & $0.484 \pm 0.21$ \\
MaskGIT (Halton) & $0.196$ & $18.20 \pm 3.50$ & $0.452 \pm 0.23$ \\
MAR-L            & $0.174$ & $20.15 \pm 4.60$ & $0.550 \pm 0.20$ \\
\textbf{Ours}    & $\mathbf{0.125}$ & $\mathbf{24.65 \pm 3.34}$ & $\mathbf{0.682 \pm 0.17}$ \\
\bottomrule
\end{tabular}
\end{table}

\subsection{Ablations}
\label{sec:ablations}

The motivation in Section~\ref{sec:motivation} makes two falsifiable predictions: future-aware positional embeddings are necessary to make the any-order objective well-posed, and the ordering curriculum is necessary to make it reachable from a pretrained initialization. We test each prediction directly.

\paragraph{Future-aware embeddings are necessary.}
\label{sec:ablation-embeddings}

Removing future-aware embeddings (Figure~\ref{fig:loss-curves}) prevents the loss from descending below a substantial floor: as the curriculum ramps in random orderings, the same observed context must explain progressively more diverse target distributions, and the loss \emph{increases} rather than decreases. This is the gradient-conflict failure mode predicted by the identifiability argument in Section~\ref{sec:method}, observed empirically. The corresponding generation quality is poor across all evaluation orders.

\paragraph{The curriculum is necessary.}
\label{sec:ablation-curriculum}

With $\alpha = 1$ from step zero (i.e., applying the well-posed any-order objective to the pretrained checkpoint without any schedule), at convergence the no-curriculum model trails our full recipe across all fidelity metrics on LlamaGen-L (Table~\ref{tab:ablation-curriculum}), while Recall is essentially unchanged, a fidelity collapse rather than a coverage collapse. The pretrained representations were eroded before order-flexible routing could form.

\begin{table}[ht!]
\centering
\caption{Effect of the ordering curriculum on LlamaGen-L. Adapting with $\alpha = 1$ from step zero (no curriculum) leaves the model in a partially adapted state across all metrics; the linear curriculum used in our main results recovers the gap.}
\label{tab:ablation-curriculum}
\small
\begin{tabular}{lcccc}
\toprule
Variant & FID $\downarrow$ & IS $\uparrow$ & Prec.\ $\uparrow$ & Rec.\ $\uparrow$ \\
\midrule
No curriculum ($\alpha = 1$ from step $0$) & $5.86$ & $191.27$ & $0.78$ & $0.51$ \\
\textbf{Ours} (linear curriculum)          & $\mathbf{4.24}$ & $\mathbf{231.62}$ & $\mathbf{0.83}$ & $\mathbf{0.52}$ \\
\bottomrule
\end{tabular}
\end{table}

\paragraph{Embedding learning rate.}
\label{sec:ablation-design}

The future-aware embedding table $E$ is zero-initialized, which means it must develop useful values from scratch even as the curriculum increases the model's reliance on it. We find empirically that scaling the learning rate of $E$ by $10\times$ relative to the rest of the network is necessary for stable convergence: at the base learning rate, the embedding develops too slowly to keep pace with the curriculum, and final any-order quality suffers. We use the $10\times$ multiplier in all main results.

\paragraph{Alternative mitigation: teacher regularization.}
\label{sec:ablation-distill}
Adding a KL term ($\lambda = 1.0$) toward the frozen raster-scan teacher on raster-ordered examples raises LlamaGen-L FID from $4.24$ to $9.62$, suggesting that matching raster-specific predictions conflicts with learning arbitrary-order conditionals.

\subsection{Training and Inference Efficiency}
\label{sec:compute}

The central practical claim of this paper is that any-order capability can be obtained at a small fraction of the cost of training such a model from scratch. Table~\ref{tab:compute} reports total training compute cost, expressed as a multiple of the corresponding raster-scan pretraining budget, alongside compatibility with the standard AR inference ecosystem.

\begin{table}[ht!]
\centering
\caption{Training and inference compatibility relative to standard AR pretraining. Our adaptation recipe achieves any-order generation at $\approx 1.25\times$ the cost of standard pretraining (LlamaGen-XL), while preserving compatibility with the AR inference ecosystem.}
\label{tab:compute}
\small
\begin{tabular}{lcccc}
\toprule
Method & Train cost & Any-order & KV cache & vLLM \\
\midrule
LlamaGen pretraining~\cite{sun2024autoregressive} (raster only)  & $1\times$           & \xmark & \cmark & \cmark \\
RandAR~\cite{pang2025randar} (from scratch)                  & $\text{3}\times$    & \cmark & \cmark & \xmark \\
RAR~\cite{yu2025randomized} (from scratch)                     & $\text{4}\times$    & \xmark & \cmark & \cmark \\
Discrete diffusion~\cite{chang2022maskgit} & $\text{3}\times$    & \cmark & \xmark & \xmark \\
\textbf{Ours} (LlamaGen-XL + adaptation) & $\mathbf{1.25\times}$ & \cmark & \cmark & \cmark \\
\bottomrule
\end{tabular}
\end{table}

Three observations follow. First, the cost of unlocking any-order generation is roughly $1.25\times$ the original pretraining run on LlamaGen-XL ($1.13$--$1.33\times$ across base models, lowest on RAR-XL, where adaptation amortizes over a longer pretraining run), compared to $\approx 3\times$ for from-scratch any-order training, with the gap widening as the base model grows. The future-aware embedding adds $(n+1) \cdot d$ parameters, $0.74$M ($0.095\%$) for LlamaGen-XL. Second, our model retains compatibility with the AR inference ecosystem (KV caching, tensor and pipeline parallelism, and vLLM~\cite{kwon2023efficient} serving through a custom model registration that supplies spatial position indices and the future-aware embedding) because the recipe leaves the standard causal single-stream transformer untouched. Discrete-diffusion any-order approaches~\cite{chang2022maskgit,weber2024maskbit} sacrifice this compatibility by construction: their non-causal attention re-attends to all unmasked positions at every refinement step, incompatible with the per-token cache reuse that powers AR inference frameworks. Third, the model admits \emph{parallel decoding} as a direct consequence of any-order training: tokens whose conditionals are near-independent given observed context can be sampled jointly within a single forward pass. On LlamaGen-XL, $9\times$ fewer forward passes cost $1.12$ FID (Table~\ref{tab:parallel-decoding}); Figure~\ref{fig:qualitativeimagenet} (bottom row) shows LlamaGen-L at $64$ steps.

\begin{table}[ht!]
\centering
\caption{Parallel decoding on LlamaGen-XL: generation steps vs.\ class-conditional FID.}
\label{tab:parallel-decoding}
\small
\begin{tabular}{lcccc}
\toprule
Generation steps & $576$ & $192$ & $128$ & $64$ \\
\midrule
FID $\downarrow$ & $3.91$ & $4.12$ & $4.81$ & $5.03$ \\
\bottomrule
\end{tabular}
\end{table}

At maximum native-setting throughput on one RTX PRO 6000, our model takes $35.6$ ms per image (batch $64$), against $56.3$ ms for MAR-L and $23$--$24$ ms for MaskGIT, whose $12$-step decoders are faster but reach worse inpainting FID; this is not a matched-FID comparison.

\section{Conclusion}
\label{sec:conclusion}

We presented a recipe for converting a pretrained raster-scan autoregressive image generator into an any-order generator through targeted adaptation. The recipe combines a future-aware positional embedding that resolves the identifiability problem inherent to causal any-order generation, and an ordering curriculum that traces a stable optimization path from the pretrained solution to a good any-order basin. Across four base models spanning two architectures and two scales (LlamaGen-L/XL, RAR-L/XL), the adapted models trade a $1.17$--$1.93$ increase in class-conditional FID relative to the pretrained checkpoints for any-order capability, outperform MaskGIT on inpainting (FID $6.21$ vs.\ $10.44$) and outpainting (FID $5.84$ vs.\ $12.17$) and are competitive with MAR-L despite never being trained on either task, and remain compatible with the standard AR inference ecosystem (KV caching, parallel decoding, vLLM via a custom model). For LlamaGen-XL, the marginal cost of adaptation is approximately $25\%$ of the original pretraining budget, against the $\approx 3\times$ required by from-scratch any-order training.

The chain rule of probability guarantees that any model that fits the joint distribution under raster-scan order is, in principle, a fit under every other ordering as well. Our results suggest that this principle is also empirically actionable: the representations a pretrained raster-scan model has already learned are not specific to scan order, and unlocking access to non-sequential conditional pathways is a much smaller intervention than learning the joint from scratch. We do not claim that causal AR should replace diffusion or flow matching; our contribution is reusing existing causal checkpoints.

\paragraph{Limitations.}
Our experiments are restricted to class-conditional image generation on ImageNet at $256 \times 256$; whether the recipe extends to text-to-image AR models or higher resolutions remains to be tested. We adapt token-based AR models specifically, and our diagnosis of the identifiability problem applies to any causal any-order generator; non-token-based variants such as MAR (which predicts continuous tokens via a diffusion head) lie outside the scope of models we adapt. Our editing evaluation covers region completion only. Finally, our random-initialization control (Section~\ref{sec:quality}) matches compute but not schedule.

\begin{ack}
Rishav Pramanik and Zhaozheng Yin have been supported by NSF grants CMMI-2246673 and CAIF-2331769, and SUNY AI Platform Award. Marco Pedersoli was
supported by the Natural Sciences and Engineering Research Council of Canada.
\end{ack}
\bibliographystyle{unsrt}
\bibliography{references}

\appendix

\section{Extended Motivation}
\label{app:motivation-extended}

This appendix expands the gradient-magnitude and gradient-variance argument summarized in Section~\ref{sec:curriculum-motivation}, and frames the ordering curriculum as a continuation method.

\paragraph{Per-permutation loss.}
Recall the any-order objective, Equation~\ref{eq:any-order-objective}. With future-aware embeddings making the per-step target identifiable, the per-permutation loss at parameters $\theta$ is
\begin{equation}
\mathcal{L}_\sigma(\theta) = -\sum_{t=1}^{n} \log p_\theta\bigl(x_{\sigma(t)} \,\big|\, x_{\sigma(<t)},\, \sigma(t+1)\bigr).
\label{eq:per-perm-loss}
\end{equation}
At the pretrained initialization $\theta^*$, this loss takes very different values depending on $\sigma$. For $\sigma$ near raster scan, $\mathcal{L}_\sigma(\theta^*)$ is small: the model is already a good fit. For random $\sigma$, $\mathcal{L}_\sigma(\theta^*)$ is large: the model has never been trained to predict in such orders.

\paragraph{Gradient magnitude.}
The gradient norms $\|\nabla \mathcal{L}_\sigma(\theta^*)\|$ inherit this imbalance: small for raster-like permutations, large for random ones. The expected gradient under $\mathrm{Uniform}(S_n)$ is therefore dominated by high-loss permutations. Even though most $\sigma \neq \sigma_{\mathrm{raster}}$ are unfamiliar to the model, they collectively contribute the bulk of the gradient signal at $\theta^*$. Early updates therefore optimize aggressively for random-permutation loss, and in doing so they rapidly modify the parameters that encode the raster-scan distribution, the very representations we wish to preserve.

\paragraph{Gradient variance.}
Different random permutations $\sigma_1, \sigma_2$ induce conditionals with different dependency structures and therefore prescribe different parameter updates. The directions $\nabla \mathcal{L}_{\sigma_1}(\theta^*)$ and $\nabla \mathcal{L}_{\sigma_2}(\theta^*)$ can disagree substantially. Stochastic gradients computed over batches of random permutations therefore have high variance, causing optimization to oscillate rather than converge and, combined with the magnitude problem, to oscillate in directions that destabilize the pretrained basin.

The combined effect is catastrophic forgetting: the network's order-flexible routing has not yet developed when its raster-scan competence is being eroded, leaving the model worse on raster scan and not yet good on anything else.

\paragraph{Curriculum as a continuation method.}
The curriculum addresses both failure modes. At small $\alpha$ in Equation~\ref{eq:curriculum}, the gradient is dominated by raster-scan terms whose magnitude is small at $\theta^*$, and the model stays close to the pretrained basin. The small contribution from random permutations is enough to begin shaping order-flexible routing without destabilizing the existing representations. As $\alpha$ increases, the model has already partially adapted, so the random-permutation loss at the current parameters is no longer catastrophically large; magnitude and variance across random permutations both decrease. This is a \emph{continuation method}: rather than leaping from $\theta^*$ to a good any-order solution in a single optimization, we trace a path through parameter space along which each successive objective is a small perturbation of the last. At every intermediate $\alpha$, the previous solution is a good initialization for the current objective, and the model's competence on prior orderings is preserved as a foundation for new ones.

\paragraph{Design-choice asides.}
We considered injecting the future-aware embedding $E_{\sigma(t+1)}$ at every transformer layer (analogous to AdaLN-style conditioning~\citep{peebles2023scalable}), but found a single input-layer addition sufficient, consistent with the view that the embedding's role is to disambiguate the prediction target rather than to participate in feature computation throughout the network.

\section{Hyperparameters}
\label{app:hyperparameters}

Table~\ref{tab:hyperparams} reports the full set of hyperparameters used to adapt each pretrained checkpoint with our recipe. All models are adapted on ImageNet from the publicly released raster-scan checkpoints at their released resolutions. The adaptation budget $T_{\mathrm{ft}}$ is set per base model to amortize against the corresponding pretraining cost (see Section~\ref{sec:compute}). The curriculum coefficient $\alpha$ ramps linearly from $0$ to $1$ over the first $S_{\mathrm{ramp}}$ epochs and remains at $1$ for the remainder of adaptation; we report $S_{\mathrm{ramp}}$ per model below.

\begin{table}[ht!]
\centering
\caption{Adaptation hyperparameters by base model. Notation matches Section~\ref{sec:method}.}
\label{tab:hyperparams}
\small
\setlength{\tabcolsep}{5pt}
\begin{tabular}{lcccc}
\toprule
& LlamaGen-L & LlamaGen-XL & RAR-L & RAR-XL \\
\midrule
\multicolumn{5}{l}{\emph{Optimization}} \\
Optimizer                                  & AdamW            & AdamW            & AdamW            & AdamW \\
$(\beta_1, \beta_2)$                       & $(0.9, 0.95)$    & $(0.9, 0.95)$    & $(0.9, 0.95)$    & $(0.9, 0.95)$ \\
Weight decay                               & $0.05$           & $0.05$           & $0.05$           & $0.05$ \\
Base learning rate                         & $1\!\times\!10^{-4}$ & $1\!\times\!10^{-4}$ & $5\!\times\!10^{-5}$ & $5\!\times\!10^{-5}$ \\
Embedding-table LR multiplier              & $10\times$       & $10\times$       & $10\times$       & $10\times$ \\
LR schedule                                & cosine decay     & cosine decay     & cosine decay     & cosine decay \\
Warmup steps                               & $0$              & $0$              & $0$              & $0$ \\
Gradient clip                              & $1.0$            & $1.0$            & $1.0$            & $1.0$ \\
Mixed precision                            & bf16             & bf16             & bf16             & bf16 \\
\midrule
\multicolumn{5}{l}{\emph{Adaptation budget}} \\
Pretraining length (epochs)                & $300$            & $100$            & $400$            & $400$ \\
$T_{\mathrm{ft}}$ (epochs)                 & $100$            & $25$             & $100$            & $50$ \\
$T_{\mathrm{ft}}$ as \% of pretraining     & $33.3\%$         & $25.0\%$         & $25.0\%$         & $12.5\%$ \\
$S_{\mathrm{ramp}}$ (epochs)               & $25$             & $10$             & $30$             & $25$ \\
Sequence length (tokens/image)             & $256$            & $576$            & $256$            & $256$ \\
\midrule
\multicolumn{5}{l}{\emph{Future-aware embedding}} \\
Embedding dim $d$                          & matches model    & matches model    & matches model    & matches model \\
Initialization                             & zero             & zero             & RAR ckpt         & RAR ckpt \\
Null embedding $E_\varnothing$             & learned          & learned          & learned          & learned \\
\midrule
\multicolumn{5}{l}{\emph{Sampling (evaluation)}} \\
Decoding                                   & top-$k$          & top-$k$          & top-$k$          & top-$k$ \\
Temperature                                & $1.0$            & $1.0$            & $1.0$            & $1.0$ \\
Classifier-free guidance scale             & $1.75$           & $1.75$           & paper schedule$^\ddagger$ & paper schedule$^\ddagger$ \\
\bottomrule
\end{tabular}
\\[2pt]
{\small $^\ddagger$For RAR-L/XL, we follow the per-step CFG schedule used by the original RAR paper~\cite{yu2025randomized} for a like-for-like comparison.}
\end{table}

The remaining configuration (architecture, tokenizer, conditioning) matches the corresponding pretraining setup of each base model and is unchanged by adaptation. For LlamaGen, the future-aware embedding adds $(n+1) \cdot d$ parameters, where $n$ is the sequence length, $d$ the hidden dimension, and the extra row is the null embedding ($0.26$M for LlamaGen-L, $0.74$M for LlamaGen-XL). RAR already contains a target-aware positional embedding, which we load from its checkpoint.

Global batch size is not fixed across runs: it depended on the GPU configuration of the node we were allocated, with the per-run setting chosen to maximize GPU utilization on that allocation. Total effective batch size was held within the range used by each base model's pretraining configuration via gradient accumulation when needed.

\section{Training and Compute Details}
\label{app:training}

\paragraph{Hardware.}
All adaptation experiments were run on internal clusters with two node types:
\begin{itemize}
    \item \textbf{H200 nodes:} $8\!\times\!$NVIDIA H200 ($144$\,GB HBM each), $192$ CPU cores per node.
    \item \textbf{RTX PRO 6000 nodes:} $4\!\times\!$NVIDIA RTX PRO 6000 Blackwell Server Edition ($96$\,GB each), $64$ CPU cores per node.
\end{itemize}
Smaller base models (LlamaGen-L, RAR-L) were adapted on the RTX PRO 6000 cluster; larger models (LlamaGen-XL, RAR-XL) used H200 nodes to accommodate the larger sequence length and per-step memory. LlamaGen-XL adaptation took $\approx 64$ H200-GPU-hours (one $8\times$H200 node for $\approx 8$ hours); we did not log per-run GPU-hours for the other models.

\paragraph{Parallelism.}
Adaptation uses standard data-parallel training; we did not use tensor or pipeline parallelism, since both base-model families fit on a single GPU at the batch sizes we use. Activation checkpointing is enabled for the XL variants. We use bfloat16 mixed precision throughout, with float32 master weights for the optimizer state.

\paragraph{Permutation sampling.}
At each training step $s$, each example in the global batch independently draws a permutation $\sigma$ from the curriculum mixture $P_{\alpha(s)}$ defined in Equation~\ref{eq:curriculum}: with probability $1 - \alpha(s)$ the raster-scan order is used, and with probability $\alpha(s)$ a fresh uniform random permutation over the $n$ image-token positions is sampled. We do not pre-sample permutations or share them across examples in a batch.

\paragraph{Evaluation protocol.}
We follow the standard ImageNet $256 \times 256$ class-conditional evaluation protocol: $50{,}000$ samples generated under the same number of decoding steps as the corresponding pretrained model (Table~\ref{tab:hyperparams}, ``Steps'' column in Table~\ref{table:main_results}), with FID, IS, and Precision/Recall computed against the standard ImageNet reference statistics. For inpainting and outpainting, masks are applied identically across methods being compared, and FID, IS, and Precision/Recall are reported over the full image (PSNR and SSIM in Table~\ref{tab:region} are computed over the generated region).

\section{Adaptation Algorithm}
\label{app:pseudocode}

Algorithm~\ref{alg:adaptation} gives the per-step adaptation loop in pseudocode. The procedure is identical to standard causal AR fine-tuning except for three additions: the per-example permutation $\sigma$ drawn from the curriculum mixture $P_{\alpha(s)}$ (line~\ref{line:sample-perm}), the spatial positions $\sigma(t)$ passed into the model's positional encoding (rotary for LlamaGen, additive learned absolute for RAR) so that positional information reflects image geometry rather than sequence step (line~\ref{line:rope-pos}), and the future-aware embedding $E_{\sigma(t+1)}$ added to each input position before the transformer forward pass (line~\ref{line:add-emb}).

\begin{algorithm}[ht]
\caption{Adaptation step for unlocking any-order generation in a pretrained AR model.}
\label{alg:adaptation}
\begin{algorithmic}[1]
\Require pretrained causal AR model $f_\theta$ with a positional encoding indexed by spatial position (e.g., RoPE or additive learned absolute)
\Require future-aware embedding table $E \in \mathbb{R}^{n \times d}$, initialized to zero (loaded from the checkpoint for RAR)
\Require null embedding $E_\varnothing \in \mathbb{R}^d$
\Require curriculum schedule $\alpha(s) = \min(s / S_{\mathrm{ramp}}, 1)$
\Require number of adaptation steps $T_{\mathrm{ft}}$
\For{$s = 1$ \textbf{to} $T_{\mathrm{ft}}$}
  \State sample minibatch of images, tokenize each into $x = (x_1, \ldots, x_n)$
  \For{each example in the minibatch}
    \State $u \sim \mathrm{Uniform}[0, 1]$
    \If{$u < \alpha(s)$} \label{line:sample-perm}
      \State $\sigma \gets$ uniform random permutation of $(1, \ldots, n)$
    \Else
      \State $\sigma \gets (1, 2, \ldots, n)$ \Comment{raster-scan}
    \EndIf
    \State reorder tokens: $\tilde{x} \gets (x_{\sigma(1)}, \ldots, x_{\sigma(n)})$
    \State spatial position vector: $\mathbf{p} \gets (\sigma(1), \sigma(2), \ldots, \sigma(n))$ \label{line:rope-pos}\Comment{passed to positional encoding}
    \State compute base input representations $h_t$ from $\tilde{x}_{<t}$ as in standard AR
    \For{$t = 1$ \textbf{to} $n$} \label{line:add-emb}
      \If{$t < n$}
        \State $\tilde{h}_t \gets h_t + E_{\sigma(t+1)}$
      \Else
        \State $\tilde{h}_t \gets h_t + E_\varnothing$ \Comment{no successor at the final step}
      \EndIf
    \EndFor
    \State predict logits $\ell_t \gets f_\theta(\tilde{h}_{1:t};\, \mathbf{p}_{1:t})$ for each $t$ \Comment{positional encoding uses spatial $\mathbf{p}$}
    \State per-example loss: $\mathcal{L}_\sigma \gets -\sum_{t=1}^{n} \log p(x_{\sigma(t)} \mid \ell_t)$
  \EndFor
  \State $\mathcal{L} \gets$ mean of per-example losses over the minibatch
  \State backpropagate; update $\theta$ at base learning rate, $E$ and $E_\varnothing$ at $10\times$ base learning rate
\EndFor
\end{algorithmic}
\end{algorithm}

\paragraph{Inference.}
Algorithm~\ref{alg:inference} describes generation under an arbitrary order $\sigma$. The order can be raster-scan (recovering the pretrained model's behavior), uniform random, or task-induced (observed tokens placed first in $\sigma$, with the remainder generated in any chosen order). Because the model is a standard causal single-stream transformer, the keys and values of generated tokens are cached and reused exactly as in standard AR inference; the future-aware embedding only affects the input representation at the current step.

\begin{algorithm}[ht]
\caption{Any-order generation with the adapted model.}
\label{alg:inference}
\begin{algorithmic}[1]
\Require adapted model $f_\theta$, embedding table $E$, null embedding $E_\varnothing$
\Require generation order $\sigma$ (a permutation of $(1, \ldots, n)$)
\Require optional observed tokens $\{x_j\}_{j \in \mathcal{O}}$ (e.g., from an inpainting mask)
\State place observed positions first in $\sigma$ if any
\State spatial position vector: $\mathbf{p} \gets (\sigma(1), \sigma(2), \ldots, \sigma(n))$ \Comment{passed to positional encoding at every step}
\For{$t = 1$ \textbf{to} $n$}
  \If{$\sigma(t) \in \mathcal{O}$}
    \State $x_{\sigma(t)} \gets$ observed value
  \Else
    \State compute base input $h_t$ from previously placed tokens $x_{\sigma(<t)}$
    \If{$t < n$}
      \State $\tilde{h}_t \gets h_t + E_{\sigma(t+1)}$
    \Else
      \State $\tilde{h}_t \gets h_t + E_\varnothing$
    \EndIf
    \State $\ell_t \gets f_\theta(\tilde{h}_{1:t};\, \mathbf{p}_{1:t})$ \Comment{positional encoding uses spatial $\mathbf{p}$; KV cache reused}
    \State $x_{\sigma(t)} \sim \mathrm{softmax}(\ell_t / \tau)$ \Comment{or top-$k$ / top-$p$}
  \EndIf
\EndFor
\State \Return image tokens $(x_1, \ldots, x_n)$, decode with the base-model VQ tokenizer
\end{algorithmic}
\end{algorithm}

\section{Broader Impacts}
\label{app:broader-impacts}

Our recipe extends the capabilities of existing pretrained class-conditional image generators rather than producing a new class of generative model. It inherits both their positive applications, such as image editing and content completion, and their misuse risks, such as misleading content generation. Because the base models are already publicly released and restricted to ImageNet classes, the recipe does not introduce new categories of risk beyond what those models already enable.

\section{Licenses for Existing Assets}
\label{app:licenses}

\paragraph{Base models.}
We adapt publicly released pretrained checkpoints from LlamaGen~\cite{sun2024autoregressive} and RAR~\cite{yu2025randomized}. LlamaGen weights are released under the MIT License via the official repository; RAR weights are released under the Apache License 2.0 via the official repository. Both license terms permit research use, redistribution, and the publication of derived models, which covers the adapted checkpoints.

\paragraph{Tokenizers.}
The VQ tokenizers we use are released alongside their corresponding base models and inherit the same licenses. We use them unmodified.

\paragraph{Dataset.}
Adaptation and evaluation use ImageNet (ILSVRC-2012)~\cite{russakovsky2015imagenet}, accessed under the standard research-use terms published by the dataset maintainers. We do not redistribute any ImageNet imagery; only metric-level results derived from generated samples are reported.

\paragraph{Software.}
We use \texttt{vLLM}~\cite{kwon2023efficient} (Apache 2.0) for inference benchmarks. All other software dependencies (PyTorch, HuggingFace utilities) are used under their standard open-source licenses.

\paragraph{Released artifacts.}
Released adapted checkpoints inherit the more restrictive of the base-model licenses they derive from and will be accompanied by a model card documenting training data provenance, intended use, and known limitations.

\end{document}